\pdfoutput=1

\documentclass[11pt]{article}

\usepackage{EMNLP2023}

\usepackage{times}
\usepackage{latexsym}

\usepackage[T1]{fontenc}

\usepackage[utf8]{inputenc}

\usepackage{microtype}

\usepackage{inconsolata}

\usepackage{listings}
\usepackage{hyperref}
\usepackage{multirow}
\usepackage{tabularx}
\usepackage{booktabs}
\usepackage{float}
\usepackage[section]{placeins}
\usepackage{caption}
\usepackage{subcaption}
\usepackage{graphicx}
\usepackage{bbm}
\usepackage{changes}

\newcommand{\sjc}[1]{\textcolor{black}{#1}}

\title{Error Detection for Text-to-SQL Semantic Parsing}
\author{
Shijie Chen \quad Ziru Chen \quad Huan Sun \quad Yu Su \\
The Ohio State University\\
  {\small \texttt{\{chen.10216, chen.8336, sun.397, su.809\}@osu.edu}} \\
  }

\begin{document}
\maketitle
\begin{abstract}

Despite remarkable progress in text-to-SQL semantic parsing in recent years, the performance of existing parsers is still far from perfect. 
Specifically, modern text-to-SQL parsers based on deep learning are often over-confident, thus casting doubt on their trustworthiness when deployed for real use.
In this paper, we propose a parser-independent error detection model for text-to-SQL semantic parsing.
Using
a language model of code
as its bedrock, we enhance our error detection model with graph neural networks that learn structural features of both natural language questions and SQL queries.
We train our model on realistic parsing errors collected from a cross-domain setting, which leads to stronger generalization ability.
Experiments with three strong text-to-SQL parsers featuring different decoding mechanisms show that our approach outperforms parser-dependent uncertainty metrics. 
Our model could also effectively improve the performance and usability of text-to-SQL semantic parsers regardless of their architectures\footnote{Our implementation is available at \url{https://github.com/OSU-NLP-Group/Text2SQL-Error-Detection}.}.

\end{abstract}
\section{Introduction}

Recent years have witnessed a renewed interest in text-to-SQL semantic parsing \citep{Bogin2019Global,Lin2020Bridging,Wang2020RATSQLa, Rubin2021SmBoP,Cao2021LGESQL,Gan2021Natural,Scholak2021PICARD,qi-etal-2022-rasat,li2022resdsql}, which allows users with a limited technical background to access databases through a natural language interface.
Although state-of-the-art semantic parsers have achieved remarkable performance on Spider \citep{Yu2018Spider}, a large-scale cross-domain text-to-SQL benchmark, their performance is still far from satisfactory for real use. 
While syntax errors can be automatically caught by SQL execution engines, detecting semantic errors in executable SQL queries can be non-trivial and time-consuming even for experts \cite{bugfixhard1, bugfixhard2}.
Therefore, an accurate error detector that can flag parsing issues and accordingly trigger error correction procedures \cite{chen-etal-2023-sqledit} can contribute to building better natural language interfaces to databases.

Researchers have proposed multiple approaches for error detection in text-to-SQL parsing. \citet{Yao2019Modelbased, Yao2020Imitation} detect errors by setting a threshold on the prediction probability or dropout-based uncertainty of the base parser. 
However, using these parser-dependent metrics requires the base parser to be calibrated, which limits their applicability.
Several interactive text-to-SQL systems detect parsing errors based on uncertain span detection \citep{Gur2018DialSQL,Li2020What,Zeng2020Photon}. 
Despite having high coverage for errors, this approach is reported to be of low precision. 
Finally, text-to-SQL re-rankers \citep{yin-neubig-2019-reranking, Kelkar2020BertrandDR, Bogin2019Global, Arcadinho2022T5QL}, which estimate the plausibility of SQL predictions, can be seen as on-the-fly error detectors. Nevertheless, existing re-rankers are trained on in-domain parsing errors, limiting their generalization ability.

In this work, we propose a generalizable and parser-independent error detection model for text-to-SQL semantic parsing.
Since syntax errors can be easily detected by an execution engine, we focus on detecting semantic errors in executable SQL predictions. 
We start developing our model with CodeBERT \citep{Feng2020CodeBERT}, a language model pre-trained on multiple programming languages. 
On top of that, we use graph neural networks to capture compositional structures in natural language questions and SQL queries to improve the performance and generalizability of our model. 
We train our model on parsing mistakes collected from a realistic cross-domain setting, which is indispensable to the model's strong generalization ability.
Furthermore, we show that our model is versatile and can be used for multiple tasks, including error detection, re-ranking, and interaction triggering. To summarize, our contributions include:
\begin{itemize}
    \item We propose the first generalizable and parser-independent error detection model for text-to-SQL parsing that is effective on multiple tasks and different parser designs without any task-specific adaptation. Our evaluations show that the proposed error detection model outperforms parser-dependent uncertainty metrics and could maintain its high performance under cross-parser evaluation settings. 
    \item
    Our work is the first comprehensive study on error detection for text-to-SQL parsing. 
    We evaluate the performance of error detection methods on both correct and incorrect SQL predictions. 
    In addition, we show through simulated interactions that a more accurate error detector could significantly improve the efficiency and usefulness of interactive text-to-SQL parsing systems.
\end{itemize}

\section{Related Work}

\subsection{Text-to-SQL Semantic Parsing}
Most existing neural text-to-SQL parsers adopt three different decoding mechanisms. The first one is sequence-to-sequence with constrained decoding, where a parser models query synthesis as a sequence generation task and prunes syntactically invalid parses during beam search. Several strong text-to-SQL parsers apply this simple idea, including BRIDGE v2 \citep{Lin2020Bridging}, PICARD \citep{Scholak2021PICARD}, and RESDSQL \cite{li2022resdsql}. Another popular decoding mechanism is grammar-based decoding \citep{Yin2017Syntactica}, where parsers first synthesize an abstract syntax tree based on a pre-defined grammar and then convert it into a SQL query. Parsers using intermediate representations, such as IR-Net \citep{Guo2019Towards} and NatSQL \citep{Gan2021Natural} also fall into this category. Grammar-based decoding ensures syntactic correctness but makes the task harder to learn due to the introduction of non-terminal syntax tree nodes. 
Different from the above autoregressive decoding strategies, SmBoP \citep{Rubin2021SmBoP} applies bottom-up decoding where a SQL query is synthesized by combining parse trees of different depths using a ranking module. 
We evaluate our model with semantic parsers using each of these three decoding strategies and show that our model is effective on all of them.

\subsection{Re-ranking for Text-to-SQL Parsing}
Noticing the sizable gap between the accuracy and beam hit rate of semantic parsers, researchers have explored building re-ranking models to bridge this gap and improve parser performance. Global-GNN \cite{Bogin2019Global} re-ranks beam predictions based on the database constants that appear in the predicted SQL query. This re-ranker is trained together with its base parser. More recently, Bertrand-DR \citep{Kelkar2020BertrandDR} and T5QL \citep{Arcadinho2022T5QL} fine-tune a pre-trained language model for re-ranking.  However, both report directly re-ranking all beams using re-ranker scores hurts performance. To get performance gain from re-ranking, Bertrand-DR only raises the rank of a prediction if its re-ranking score is higher than the preceding one by a threshold. T5QL combines re-ranking score and prediction score by a weighted sum. Both approaches require tuning hyper-parameters. In contrast, when directly using the proposed parser-independent error detection model as re-rankers, we observe performance improvement on 
NatSQL without any constraint, showing that our approach is more generalizable and robust.

\begin{table*}[!htbp]
  \small
  \centering
    \begin{tabular}{l c c c c c c c c c}
      \toprule
        \multirow{3}{*}{Parser} &\multicolumn{3}{c}{Train} &\multicolumn{3}{c}{Development} & \multicolumn{3}{c}{Test}\\
      \cmidrule(lr){2-4} \cmidrule(lr){5-7} \cmidrule(lr){8-10}
         & \#Beam & Hit & Miss & \#Beam & Hit& Miss & \#Beam & Hit& Miss \\
      \midrule
          SmBoP  & 5322 & 6062/1.4 & 12937/2.6 & 1416 & 1864/1.3 & 3159/2.5 &  989 & 1324/1.3 & 1498/1.5 \\
          \sjc{RESDSQL}  & 5506 & 7329/1.3 & 11817/2.2 &  1470 & 2210/1.5 & 2835/1.9  & 1033 & 1380/1.3 & 1673/1.6 \\
          NatSQL  & 5398 & 7095/1.3 & 13443/2.5 &  1474 & 2207/1.5 & 3522/2.4 & 1030 & 1582/1.5 & 2584/2.5 \\
      \bottomrule
      \end{tabular}
      \caption{\label{tab:data_statistics}
      Statistics of the data collected from three base parsers. \#Beam: number of beams that have executable predictions. For beam hit and misses, we report total/average\_per\_beam.}
  \end{table*}
\subsection{Interactive Text-to-SQL Parsing Systems}
Interactive text-to-SQL parsing systems improve the usability of text-to-SQL semantic parsers by correcting potential errors in the initial SQL prediction through interactive user feedback. 
MISP \citep{Yao2019Modelbased,Yao2020Imitation} initiates interactions by setting a confidence threshold for the base parser's prediction probability. 
While this approach is intuitive, it requires the base parser to be well-calibrated when decoding, which does not hold for most modern parsers using deep neural networks. In addition, this design can hardly accommodate some recent parsers, such as SmBoP \citep{Rubin2021SmBoP}, whose bottom-up decoding mechanism does not model the distribution over the output space. 
Several other interactive frameworks \citep{Gur2018DialSQL,Li2020What,Zeng2020Photon} trigger interactions when an incorrect or uncertain span is detected in the input question or predicted SQL query. 
While these approaches have high coverage for parsing errors, they tend to trigger unnecessary interactions for correct initial predictions.
For example, PIIA \citep{Li2020What} triggers interactions on 98\% of the questions on Spider's development set when its base parser has an accuracy of 49\%. 
Compared to these methods, the proposed method strikes a better balance between performance and efficiency, and thus could improve the user experience of interactive text-to-SQL parsing systems.

\section{Parser-independent Error Detection}

\subsection{Problem Formulation}
Given a question $X=\{x_1, x_2, \cdots, x_m\}$ and a SQL query $\hat y = \{\hat y_1, \hat y_2, \cdots, \hat y_n\}$ predicted by a text-to-SQL parser, the error detection model estimates the probability of $\hat y$ being correct, denoted by $s$:
\[s=p(\hat y = y^*\vert X, \hat y)\]
We perform error detection and action triggering by setting a threshold for $s$. For re-ranking, we directly use $s$ as the ranking score without modification.

\subsection{Cross-domain Error Collection}
\label{Sec:data-collection}

We consider two factors that could lead to text-to-SQL parsing errors: \textit{insufficient training data} and the \textit{cross-domain generalization gap}.
To simulate such errors, we collect data from weak versions of base parsers in a cross-domain setting.
More specifically, we split the Spider training set into two equal-sized subsets by databases and train the base parser on each subset. Then we perform inference on the complementary subset and collect beam predictions as data for error detection. 
We keep executable SQL queries and label them based on execution accuracy.
We use a fixed version of Spider's official evaluation script (Appendix \ref{sec:appendix_fixes_spider}) and keep up to five parser predictions for each question after deduplication.  
The collected samples are divided into training and development sets by an 80:20 ratio according to databases as well. 
In this way, we get high-quality training data for our error detection model in a setting that approximates the real cross-domain testing environment. 
For testing, we train each base parser on the full Spider training set and collect executable beam predictions on the Spider development set.
Beams with un-executable top predictions are skipped. We report the number of beams, total number of question-SQL pairs, and average number of such pairs per beam for each split in Table \ref{tab:data_statistics}. 
Following existing literature \cite{Kelkar2020BertrandDR}, we refer to correct SQL queries in the beam as beam hits, and incorrect ones as beam misses.

\sjc{We choose three base parsers with different decoder architectures, namely SmBoP, RESDSQL\footnote{We use the RESDSQL-large + NatSQL configuration.}, and NatSQL.
We notice that the strongest base parser, RESDSQL, generates the most beam hits on the train and development splits. However, it also generates the least executable beam misses. This might be caused by its unconstrained decoder which often produces unexecutable SQL queries.}
NatSQL and SmBoP take into account grammatical constraints of SQL during decoding and thus could generate more executable queries than \sjc{RESDSQL}.
Table \ref{tab:data_statistics} also shows that \sjc{all three base parsers} produce a similar amount of beam hits and beam misses on the train and development splits. However, the number of executable beam misses generated by SmBoP \sjc{and RESDSQL} on the test split is noticeably lower, while the behavior of NatSQL is more consistent. 

\subsection{Model Architecture}

\begin{figure}[h]
  \centering
    \includegraphics[width=0.5\textwidth]{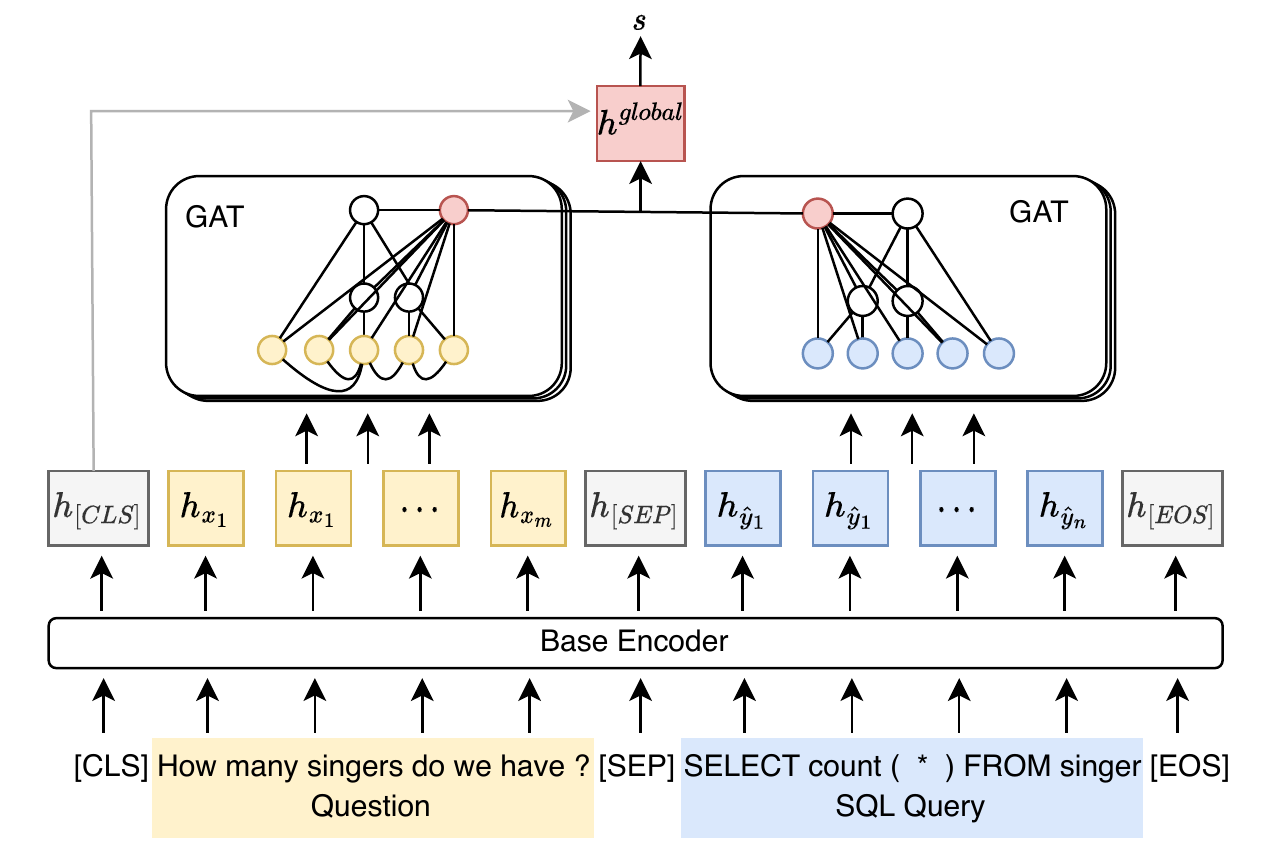}
    \caption{\label{fig:model_arch}
    Architecture of our error detection models.}
\end{figure}

Figure \ref{fig:model_arch} illustrates the architecture of the proposed error detection models.
We use CodeBERT \citep{Feng2020CodeBERT} as our base encoder to jointly encode the input question and SQL query. Following CodeBERT's input construction during pre-training, we concatenate questions and SQL queries with special tokens, namely $[CLS], x_1, \cdots, x_m,[SEP],\hat y_1,\cdots,\hat y_n,[EOS]$ as input and obtain their contextualized representations $h_X$ and $h_{\hat y}$. We only use question and SQL as input since we found in preliminary experiments that adding database schema information (table and column names) in the input hurts performance.

In light of the compositional nature of questions and SQL queries, we propose to model their structural features via graph neural networks. 
For natural language questions, we obtain their dependency parse trees and constituency parse trees from Stanza \citep{Qi2020Stanza} and merge them together. 
This is possible since edges in dependency parse trees are between two actual tokens, which corresponds to leaf nodes in constituency parse trees.
For SQL queries, we extract their abstract syntax trees via Antlr4.\footnote{Antlr: \url{https://www.antlr.org/}, more details in Appendix \ref{sec:appendix_sqlite_grammar}.} 
To make the input graphs more compact and lower the risk of overfitting, 
we further simplify the parse trees 
by removing non-terminal nodes that only have one child in a top-down order. 
Additionally, for SQL queries, we remove the subtrees for join constraints which do not carry much semantic information but are often quite long. 
At last, we add sequential edges connecting the leaf nodes in the parse trees by their order in the original questions and SQL queries to preserve natural ordering features during graph learning.

We initialize the representations of parse tree leaf nodes with CodeBERT's contextualized representations and randomly initialize representations of other nodes according to their types in the parse tree. The two input graphs are encoded by two separate 3-layer graph attention networks \citep{Brody2022How}. Then we obtain the global representation of each graph via average pooling and concatenate them to get an aggregated global representation: $$h^{global}=[h^{global}_{X}; h^{global}_{\hat y}]$$
We denote models with graph encoders as CodeBERT+GAT in Section \ref{sec:exp}.
When simply fine-tuning CodeBERT, $h^{global}=h_{[CLS]}$.

Finally, a 2-layer feed-forward neural network with $\mathrm{tanh}$ activation is used to score the aggregated representation $v$. The score $s$ for each input question-SQL pair is: 
$$s = p(\hat y = y^* \vert X, \hat y) = \sigma(\textrm{FFN}(h^{global}))$$
where $y^*$ is the gold SQL query and $\sigma$ represents the sigmoid function. We train our model by minimizing a binary cross entropy loss:
$$\mathcal{L} = \mathbbm{1}_{\hat y = y^*}\cdot \log s +(1- \mathbbm{1}_{\hat y = y^*})\cdot\log(1-s)$$
During training, we supply the model with samples from $K$ beams at each step, where $K$ is the batch size.

\section{Experiments}
\label{sec:exp}
In this section, we first evaluate the performance (Section \ref{sec:exp_ed}) and generalization ability (Section \ref{Sec:cross-parser-generalization}) of our error detection model on the binary error detection task. Then we investigate our model's effectiveness when used for re-ranking (Section \ref{sec:exp_reranking}) and action triggering (Section \ref{sec:exp_isp}).

\begin{table*}[!htbp]
  \small
  \centering
  \begin{tabular}{llcccccccc}
    \toprule
    \multirow{3}{*}{Parser}&\multirow{3}{*}{Model} & \multicolumn{3}{c}{Positive}&\multicolumn{3}{c}{Negative}&\multirow{3}{*}{Acc}&\multirow{3}{*}{AUC}\\
    \cmidrule(lr){3-5} \cmidrule(lr){6-8}
      &  & Precision & Recall & F1 & Precision & Recall& F1 &  & \\
    \midrule
    \multirow{4}{*}{SmBoP}& SmBoP$^p$  & 80.8  & \textbf{94.5} & 86.6  & 52.1 & 18.9  & 23.9  & 77.4  &  67.0  \\
    & SmBoP$^s$  & 81.5 & 91.9  & 85.7  & 56.6 & 25.3 & 29.4  & 76.9 & 79.2  \\
    \cmidrule(lr){2-10}
    & CodeBERT & 82.9  & 92.6 & 86.7  & \textbf{60.8} & 33.9  & 36.3 & 78.3 & 80.8 \\
    & CodeBERT+GAT & \textbf{85.0}  & 90.6 & \textbf{87.2} & 56.7 & \textbf{44.4}  & \textbf{46.4} & \textbf{79.8} & \textbf{81.7} \\
    \midrule
    
    \multirow{4}{*}{\sjc{RESDSQL}}&RESDSQL$^p$  & 79.5  & 93.9 & 85.5 & 55.1 & 15.1  & 19.7  & 75.6  &  76.2  \\
    & RESDSQL$^s$  & 80.1 & \textbf{95.2} & 86.5 & 54.2 & 16.2 & 21.6  & 77.5 & 76.7  \\
    \cmidrule(lr){2-10}
    & CodeBERT & 83.1 & 94.8 & \textbf{88.3}  & 61.2 & 32.1 & 41.0 & 81.0 & \textbf{80.7} \\
    & CodeBERT+GAT & \textbf{83.8}  & 93.8 & \textbf{88.3} & \textbf{61.7} & \textbf{37.0}  & \textbf{45.2} & \textbf{81.2} & \textbf{80.7} \\
    
    \midrule
    \multirow{4}{*}{NatSQL}&NatSQL$^p$  & 78.1  & \textbf{93.2} & 84.6 & 67.3 & 36.1  & 45.4  & 76.3  &  79.2  \\
    & NatSQL$^s$  & 77.0 & 91.4 & 83.0 & 62.8 & 33.1 & 40.3  & 74.0 & 76.2  \\
    \cmidrule(lr){2-10}
    & CodeBERT & 84.6  & 90.8 & \textbf{87.3}  & \textbf{72.3} & 60.5 & 64.6 & \textbf{81.8} & 86.5 \\
    & CodeBERT+GAT & \textbf{86.6}  & 87.4 & 86.8 & 68.5 & \textbf{68.1}  & \textbf{67.0} &
    81.7 & \textbf{86.9} \\
    \bottomrule
  \end{tabular}  
  \caption{\label{tab:ed_perf}
  Error detection performance with three base parsers on Spider's development set. We highlight the best performance with each parser in bold.}
\end{table*}
\subsection{Experiment Setup}
\paragraph{Baseline Methods} We compare our parser-independent error detectors with parser-dependent uncertainty metrics, including prediction probability and dropout-based uncertainty. Since SmBoP \citep{Rubin2021SmBoP} uses bottom-up decoding which separately scores and ranks each candidate prediction, we deduplicate SmBoP's beam predictions by keeping the maximum score and perform softmax on the deduplicated beam to get a probability distribution over candidate predictions, which can be seen as a reasonable approximation to its confidence. 
RESDSQL \citep{li2022resdsql}
and NatSQL \citep{Gan2021Natural} use autoregressive decoders, and we directly use the log probability of its prediction as its confidence score. 
Probability-based methods are denoted by superscript $p$.
In terms of dropout-based uncertainty, we follow MISP \citep{Yao2019Modelbased} and measure the standard deviation of the scores (SmBoP) or log probability (\sjc{RESDSQL} and NatSQL) of the top-ranked prediction in 10 passes. 
Dropout-based uncertainty is denoted by superscript $s$.

\paragraph{Evaluation Metrics} 
We first evaluate our model on the error detection task. After that, we test performance when it is used for re-ranking and action triggering.

For error detection, we report precision, recall, and F1 scores for each method on both positive and negative samples.
However, these metrics depend on the threshold used. 
To more comprehensively evaluate the overall discriminative ability of each method, we present the area under the receiver operating characteristic curve (AUC), which is not affected by the choice of threshold.
We apply 5-fold cross-validation and report performance using the threshold that maximizes the accuracy of each method. Test samples are partitioned by databases.

For the re-ranking task, we evaluate on the final beam predictions of fully trained base parsers on Spider's development set and report top-1 accuracy. 

For action triggering, we evaluate system performance under two settings: answer triggering and interaction triggering. In answer triggering, we measure system answer precision when answering different numbers of questions. In interaction triggering, we measure system accuracy using different numbers of interactions.  

Error detection and re-ranking results 
are average performance over 3 different random seeds. For action triggering, we evaluate checkpoints with the highest accuracy on the development split of our collected data.

\paragraph{Implementation} Our models are trained with a batch size of 16 and are optimized by the AdamW \citep{Loshchilov2019Decoupled} optimizer with default parameters. Training lasts 20 epochs with a learning rate of 3e-5 following a linear decay schedule with 10\% warm-up steps. 
All models are trained on an NVIDIA RTX A6000 GPU.

\subsection{Results}

\begin{table*}[htbp]
  \small
  \centering
  \begin{tabular}{lllcccccccc}
    \toprule
    \multirow{3}{*}{Target} & \multirow{3}{*}{Source}& \multirow{3}{*}{Model} & \multicolumn{3}{c}{Positive}&\multicolumn{3}{c}{Negative}&\multirow{3}{*}{Acc}&\multirow{3}{*}{AUC}\\
    \cmidrule(lr){4-6} \cmidrule(lr){7-9}
        &&& Precision & Recall & F1 & Precision & Recall& F1 &  & \\
    \midrule
    \multirow{7}{*}{SmBoP} & \multirow{2}{*}{-}& SmBoP$^p$  & 80.8  & 94.5 & 86.6  & 52.1 & 18.9  & 23.9  & 77.4  &  67.0  \\
    && SmBoP$^s$  & 81.5 & 91.9  & 85.7  & 56.6 & 25.3 & 29.4  & 76.9 & \textbf{79.2}  \\
    \cmidrule(lr){2-11}
    &\multirow{2}{*}{\sjc{RESDSQL}} & CodeBERT & 81.9 & 94.0 & 87.1  & 54.8 & 25.3  & 31.7 & 78.8 & 77.1 \\
    &&CodeBERT+GAT & 81.3  & 95.0 & 87.2 & 47.4 & 21.7  & 28.0  & 78.8 & 77.1 \\

    \cmidrule(lr){2-11}
    & \multirow{2}{*}{NatSQL} &  CodeBERT & 81.9  & \textbf{95.2} & \textbf{87.7}  & \textbf{58.2} & 23.4  & 30.8 & 79.6 & 75.9 \\
    &&CodeBERT+GAT & \textbf{83.1}  & 93.3 & 87.6 & 56.0 & \textbf{31.5}  & \textbf{38.1} & \textbf{79.9}& 78.2 \\
    \midrule

    \multirow{7}{*}{\sjc{RESDSQL}} &\multirow{2}{*}{-}&RESDSQL$^p$  & 79.5  & 93.9 & 85.5 & 55.1 & 15.1  & 19.7  & 75.6  &  76.2  \\

    && RESDSQL$^s$  & 80.1 & 95.2 & 86.5 & 54.2 & 16.2 & 21.6  & 77.5 & 76.7  \\
    
    \cmidrule(lr){2-11}
    & \multirow{2}{*}{SmBoP} & CodeBERT & 82.1 & \textbf{94.6} & 87.4  & 65.3 & 29.5  & 37.1 & 79.4 & 80.8 \\
    &&CodeBERT+GAT & 82.9  & 94.4 & \textbf{87.9} & \textbf{66.0} & 33.5  & 42.4 & \textbf{80.3} & 80.7 \\
    \cmidrule(lr){2-11}
    & \multirow{2}{*}{NatSQL} & CodeBERT & \textbf{84.0}  & 91.6 & 87.3  & 55.9 & \textbf{40.0}  & \textbf{44.7} & 79.9 & 80.3 \\
    && CodeBERT+GAT & 82.7 & 92.5 & 87.0 & 54.3 & 32.9 & 38.7 & 78.9 & \textbf{81.0} \\
    

    
    \midrule
    \multirow{7}{*}{NatSQL} & \multirow{2}{*}{-}&NatSQL$^p$  & 78.1  & \textbf{93.2} & 84.6 & \textbf{67.3} & 36.1  & 45.4  & 76.3  &  79.2  \\
    && NatSQL$^s$  & 77.0 & 91.4 & 83.0 & 62.8 & 33.1 & 40.3  & 74.0 & 76.2  \\
    \cmidrule(lr){2-11}
    &\multirow{2}{*}{SmBoP} &  CodeBERT & \textbf{83.7}  & 86.1 & 84.0  & 63.5 & \textbf{61.1}  & 58.5 & 77.0 & 85.2 \\
    && CodeBERT+GAT & 83.5  & 87.2 & \textbf{84.7} & 65.1 & 59.5  & \textbf{59.9} & \textbf{78.2} & \textbf{85.7} \\
    \cmidrule(lr){2-11}
    & \multirow{2}{*}{\sjc{RESDSQL}} & CodeBERT & 82.3 & 87.7 & 84.5 & 63.8 & 54.7  & 57.3 & 77.7 & 83.6 \\
    && CodeBERT+GAT & 82.9 & 86.7 & 84.4  &  63.7 & 56.9  & 58.7 & 77.9 & 83.6 \\
    \bottomrule
  \end{tabular}
  \caption{\label{tab:generalization}
  Cross-parser generalization performance with three base parsers on Spider's development set. We highlight the best performance with each target parser in bold.}
\end{table*}
\subsubsection{Error Detection}

\label{sec:exp_ed}

To evaluate error detection methods in a realistic setting, we use final SQL predictions made by SmBoP,
RESDSQL, and NatSQL on Spider's development set that are executable as test datasets. 
As shown in Table \ref{tab:ed_perf}, the dropout-based uncertainty measure significantly outperforms the approximate confidence measure on negative samples with SmBoP (29.4 vs 23.9 in negative F1) \sjc{and RESDSQL (21.6 vs 19.7 in negative F1)}. However, we notice the opposite with
NatSQL, which is consistent with the observation of MISP \citep{Yao2019Modelbased} that is based on a parser with LSTM-based decoder as well. 
Nonetheless, the dropout-based uncertainty measure is still indicative of \sjc{NatSQL's performance}.
We also notice that parser-dependent metrics exhibit high recall and low precision on positive samples, showing that the three parsers, despite using different decoding strategies, are over-confident in their predictions.

On all three parsers, our proposed error detector significantly outperforms the two parser-dependent uncertainty metrics, especially on negative samples. 
With the added structural features, CodeBERT+GAT further improves overall error detection performance, especially in recall on incorrect predictions (\sjc{7.7}\% absolute improvement on average), which indicates structural features could help the model learn more generalizable error patterns. 
We also find that the advantage of CodeBERT+GAT mainly comes from its higher performance on \textit{hard} and \textit{extra hard} questions (Appendix \ref{sec:appendix_perf_difficulty}).
Compared to parser-dependent metrics, our model yields the largest performance gain in both accuracy and AUC with NatSQL and reasonable gains with the other two parsers, possibly due to the higher quality of its training data and better behavior consistency on the test split. 

\subsubsection{Cross-parser Generalization}
\label{Sec:cross-parser-generalization}

We evaluate our models' cross-parser generalization ability by training error detectors on data collected from one parser and testing on the other two following the same 5-fold cross-validation setting. 
Table \ref{tab:generalization} summarizes cross-parser transfer performance on each parser. Even in this setting, our error detectors could still outperform parser-dependent metrics except for SmBoP, where our models fall short slightly in AUC.


\sjc{We observe that error detection models trained with data generated from SmBoP performs the best on both RESDSQL and NatSQL. Meanwhile, models trained with NatSQL performs better than models trained with RESDSQL on SmBoP. We attribute this to the diversity and coverage of training examples.}
We found that the errors generated by \sjc{RESDSQL}
and NatSQL, two autoregressive parsers, are more likely to share prefixes and differ in simple operations, such as the choice of columns, aggregation functions, or logic operators (examples in Appendix \ref{sec:appendix_beam_examples}). 
In contrast, the bottom-up decoder of SmBoP generates more diverse errors with complex structures, such as subqueries and set operations. The higher diversity of SmBoP's predictions increases the coverage of the data collected from it, which contributes to the stronger generalization ability of the corresponding error detectors.


\begin{figure*}[!htbp]
  \centering
  \includegraphics[width=\textwidth]{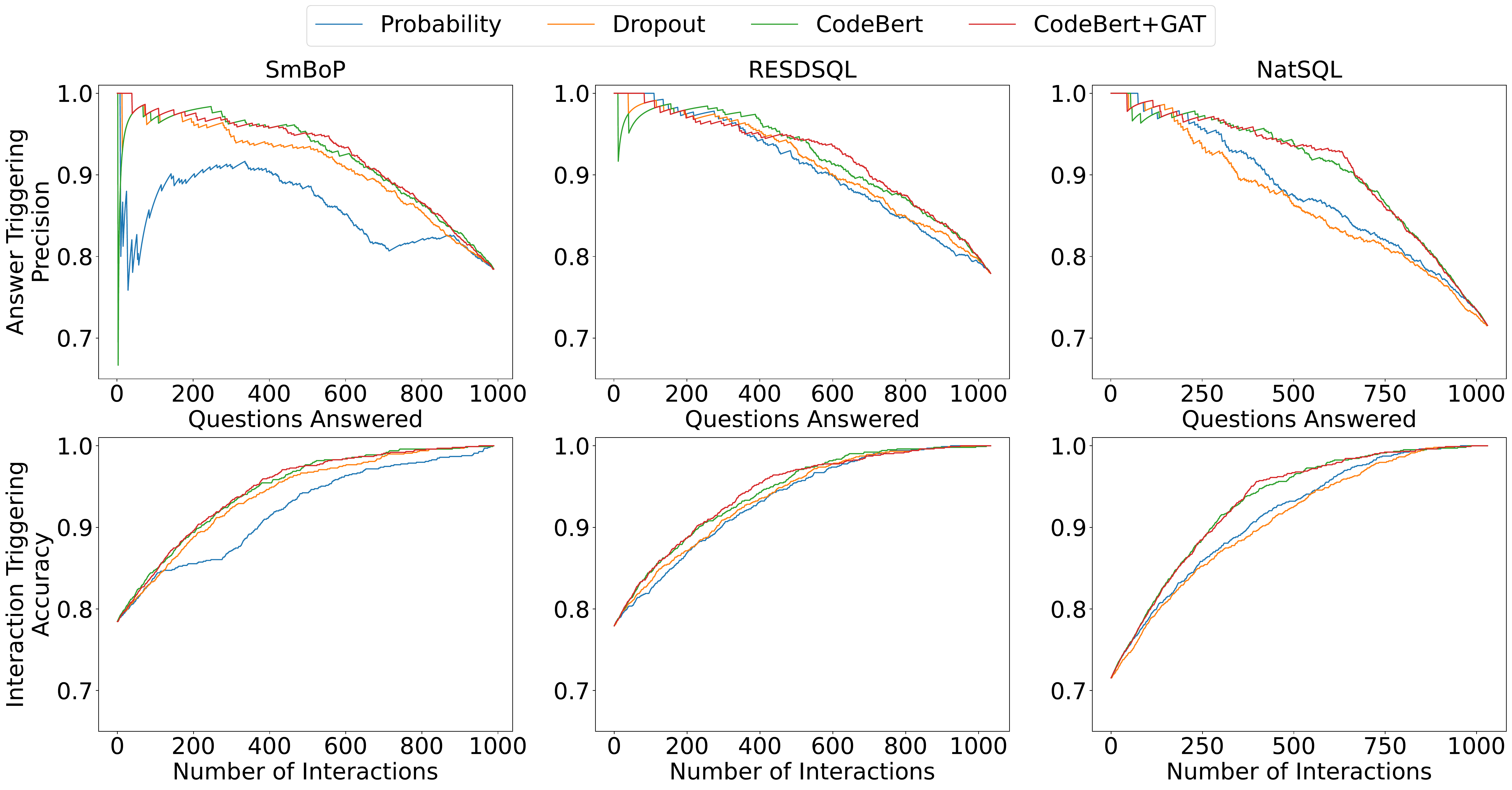}
  \caption{\label{fig:evaluation_triggering}
  Performance in simulated interactive semantic parsing with three base parsers.}
\end{figure*}

\subsubsection{Re-ranking}
\label{sec:exp_reranking}

\begin{table}[htbp]
  \small
  \centering
  \begin{tabular}[!htbp]{lccc}
    \toprule
    Re-ranker & SmBoP & \sjc{RESDSQL} & NatSQL \\
    \midrule
    N/A & \textbf{75.1} & \textbf{77.9} & 71.3 \\
    \midrule
    RR & 72.2 & 75.7& 72.5 \\
    ED + RR & 73.7 & 77.8 & \textbf{73.9} \\
    \midrule
    Beam Hit & 80.5 & 86.6 & 81.1 \\
    \bottomrule
  \end{tabular}
  \caption{\label{tab:reranking} Execution accuracy with re-ranking using the CodeBERT + GAT model. RR: Re-ranking all beams; ED+RR: Re-ranking beams after error detection.}
\end{table}


We evaluate the re-ranking performance of our error detection models in two settings. In re-ranking-all (RR),  we re-rank all beams based on the score assigned by the error detector. In error detection then re-ranking (ED+RR), we only re-rank the beams whose top-ranked prediction has a score below a given threshold. For simplicity, we use a decision threshold of $0.5$ for error detection.

As shown in Table \ref{tab:reranking}, our error detectors can improve the performance of 
NatSQL in both settings without training on any re-ranking supervision. Compared with existing re-rankers, our model does not need extra hyper-parameters for performance gain, even in the re-ranking-all setting. \sjc{However, re-ranking does not improve the performance of RESDSQL, the strongest parser among the three. 
Further, it \sjc{noticeably} hurts the performance of SmBoP. We attribute this to SmBoP's larger train-test discrepancy due to the bottom-up nature of its decoder.}
As discussed in Section \ref{Sec:data-collection} and Section \ref{Sec:cross-parser-generalization}, SmBoP produces more diverse beam predictions, but its behavior is less consistent on the test split. While the diversity benefits the quality of data for training error detectors, the inconsistency makes re-ranking on the test split harder. 
Although SmBoP is the strongest parser among the three, state-of-the-art text-to-SQL parsers predominantly use autoregressive decoders. 
Therefore, we still expect our approach to be generally applicable.
We also perform 0-shot re-ranking evaluation on the more challenging KaggleDBQA \cite{lpr2021kaggledbqa} dataset (Appendix \ref{sec:appendix_kaggledbqa}) \sjc{by training error detection models with another parser BRIDGE v2}. CodeBERT+GAT improves BRIDGE v2's accuracy from 20.5\% to 21.8\%, showing good generalization to unseen datasets.

\subsubsection{Action Triggering in Interactive Systems}
\label{sec:exp_isp}
In this section, we evaluate the potential gain of using our error detection model as an answer trigger and interaction trigger in interactive semantic parsing systems.

\paragraph{Answer triggering} When using error detectors for answer triggering, the interactive semantic parsing system restrain from answering the user's question when an error is detected. The upper half of Figure \ref{fig:evaluation_triggering} demonstrates the change of precision when varying the decision threshold. 
In general, a high threshold $p$ (or lower $s$) reduces the number of questions answered for higher precision. Conversely, a lower $p$ (or higher $s$) encourages the system to answer more questions at the cost of making more mistakes.

Because of the high precision on positive samples, the proposed error detectors outperform both baseline methods and allow the system to answer more questions at higher precision. As shown by Table \ref{tab:answer_triggering}, when maintaining a precision of 95\%, our error detectors allow the system to answer \sjc{76\%} to 175\% more questions compared to parser-dependent metrics.


\begin{table}[htbp]
  \small
  \centering
  \begin{tabular}[!htbp]{lccc}
    \toprule
    Model & SmBoP & \sjc{RESDSQL} & NatSQL \\
    \midrule
    Probability & 8 & 365 & 302 \\
    Dropout & 295 & 409 & 213 \\
    \midrule
    CodeBERT & 498 & 455 & \textbf{441} \\
    CodeBERT+GAT & \textbf{520} & \textbf{462} & 395 \\
    \bottomrule
  \end{tabular}
  \caption{\label{tab:answer_triggering} The number of questions each parser could answer when maintaining a precision of 95\%.}
\end{table}

\begin{table*}[!htbp]
  \small
  \centering
  \begin{tabular}{ccccccccc}
    \toprule
    \multirow{3}{*}{Graph}&\multirow{3}{*}{Data Source} & \multicolumn{3}{c}{NatSQL}&\multicolumn{2}{c}{\sjc{RESDSQL}}&\multicolumn{2}{c}{SmBoP}\\
    \cmidrule(lr){3-5} \cmidrule(lr){6-7}\cmidrule(lr){8-9}
      &  & Acc & AUC & RR & Acc & AUC&Acc & AUC \\
    \midrule
    \multicolumn{2}{c}{NatSQL$^p$} & 76.3 & 79.2 & 71.3 &-&-&-&-\\
    \midrule
    Simplified & in-domain & 79.0 & 83.8 & 71.2 & 78.4 & 77.0 & 78.5 & 71.3\\
    Simplified & cross-domain & \textbf{81.7} & 86.9 & \textbf{72.5} & 78.9 & \textbf{81.0} &\textbf{79.9}&\textbf{78.2}\\
    Original & cross-domain & \textbf{81.7} & \textbf{87.2} & 71.9 & \textbf{80.0} & 80.8 &79.5&76.0\\
    \bottomrule
  \end{tabular}  
  \caption{\label{tab:ablation}
  Ablation results using the CodeBERT + GAT model trained on data collected from NatSQL. We report accuracy, AUC, re-ranking-all (RR) performance on NatSQL's test split as in-domain evaluation and report accuracy and AUC when tested on SmBoP and RESDSQL as generalization evaluation. NatSQL$^p$ is included for reference.}
\end{table*}


\paragraph{Interaction triggering}


\begin{table}[htbp]
  \small
  \centering
  \begin{tabular}[!htbp]{lccc}
    \toprule
    Model & SmBoP & \sjc{RESDSQL}  & NatSQL \\
    \midrule
    Probability & 535 & 482 & 573 \\
    Dropout & 412 & 465 & 597 \\
    \midrule
    CodeBERT & 370 & 437 & 421 \\
    CodeBERT+GAT & \textbf{356} & \textbf{382} & \textbf{384} \\
    \bottomrule
  \end{tabular}
  \caption{\label{tab:interaction_triggering} The number of interactions each parser needs with interaction triggering to reach an accuracy of 95\%.}
\end{table}

We simulate the potential gain of more accurate interaction triggers by assuming oracle error correction interactions, where any detected error would be fixed through interactions with users. 
Ideally, we would want to get higher system accuracy with fewer interactions. 
The lower half of Figure \ref{fig:evaluation_triggering} illustrates the change of accuracy at different interaction budgets.

Our parser-independent models consistently improve upon parser-dependent metrics, resulting in more efficient interactive semantic parsing systems. Due to higher precision and recall on erroneous base predictions, systems using our models could correct more errors and avoid unnecessary interactions.
As shown by Table \ref{tab:interaction_triggering}, depending on the base parser, our model brings a \sjc{16\%} to 33\% reduction to the number of interactions required for reaching an accuracy of 95\%.

\subsection{Ablation}
\label{sec:ablation}
We perform ablation studies on the impact of cross-domain error collection and graph learning using the CodeBERT+GAT model.
We report the models' accuracy, AUC, and re-ranking performance in the re-rank all setting (RR) on the test split of NatSQL. We also test the models on \sjc{RESDSQL}
 and SmBoP to evaluate their generalization ability.

\paragraph{Cross-domain error collection} We train a NatSQL model using the full Spider training set and perform inference on the same data set to get its beam predictions. Then we create training data for error detection following the procedure described in Section \ref{Sec:data-collection}. In this way, we collect in-domain parsing errors in the same setting as Bertrand-DR and T5QL. As shown by Table \ref{tab:ablation}, the error detector trained on in-domain errors significantly underperforms the one trained on cross-domain errors. The performance of NatSQL deteriorates after re-ranking, which is consistent with the findings of previous re-rankers. Thus, we conclude that collecting high-quality parsing errors in a realistic cross-domain setting is critical to building an accurate and generalizable error detector. 

\paragraph{Simplified graph input} In this setting, we do not simplify constituency parse trees and SQL abstract syntax trees when constructing input graphs for graph neural networks. Table \ref{tab:ablation} shows that the model without graph simplification slightly outperforms the one using simplified graphs in AUC. Despite that, its re-ranking and cross-parser generalization performance are lower. We hypothesize that graph simplification could maintain important structural features of the input and improve the model's generalization ability by alleviating overfitting during training. 

\section{Conclusion}
In this work, we propose the first generalizable parser-independent error detection model for text-to-SQL semantic parsing. 
Through learning compositional structures in natural language and SQL queries, the proposed model significantly outperforms parser-dependent uncertainty metrics and could generalize well to unseen parsers.
We further demonstrate the versatility of our approach in error detection, re-ranking, and action triggering through a case study with three state-of-the-art text-to-SQL parsers featuring different decoding mechanisms. 

Our experiments highlight the important role of structural features and cross-domain training data in building strong and generalizable error detectors for semantic parsing. Potential future work includes (1) developing more advanced architectures to better evaluate the semantic correctness of synthesized SQL queries, (2) exploring data synthesis strategies to automatically create high-quality training data for error detection models.

\section*{Limitations}

This work is the first attempt towards building a versatile error detector for text-to-SQL semantic parsing. 
Although our model is parser-independent, the current data collection process depends on the choice of base parsers. As a result, the collected data may inherit certain biases in the base parsers. Our experiments show that data collected from stronger base parsers helps the model to generalize to weaker parsers. However, how to collect high-quality training data for error detection with stronger base parsers like SmBoP remains an open problem.
A promising future direction may be developing a comprehensive data synthesis approach to improve the quality of training data. Grappa \citep{Yu2021Grappa} uses context-free grammar to synthesize SQL queries for pre-training Transformer encoders for text-to-SQL parsing. This approach could be adapted to generate syntactically correct but semantically incorrect SQL queries in a controllable way.

Another major limitation is that our current model does not consider database schema information. Since SQL queries are grounded in databases, in principle database schema (tables, columns, and foreign-key relationships) should be an important part of error detection. The common practice in text-to-SQL semantic parsing is to linearize the database schema and concatenate all table and column names to the input to the Transformer encoder. However, our preliminary experiments show that this operation actually hurts the error detection performance. 
A similar observation is also reported by \citet{Kelkar2020BertrandDR}. 
Nevertheless, our approach performs strongly for error detection as it can still effectively capture semantic errors that are free from schema linking mistakes. This can be explained by the high column mention rate in Spider \citep{Pi2022Towards}. Future work could develop more effective entity linking mechanisms to extend our model to more challenging testing environments where schema linking errors are more common.

 \section*{Acknowledgements}
 We would like to thank colleagues from the OSU NLP group for their thoughtful comments. 
 This research was sponsored in part by a sponsored research award by Cisco Research, NSF IIS-1815674, NSF CAREER \#1942980, NSF OAC-2112606, and Ohio Supercomputer Center \citep{OhioSupercomputerCenter1987}. 
 The views and conclusions contained herein are those of the authors and should not be interpreted as representing the official policies, either expressed or implied, of the U.S. government.
 The U.S. Government is authorized to reproduce and distribute reprints for Government purposes notwithstanding any copyright notice herein.

\bibliography{ref}

\begin{thebibliography}{31}
\expandafter\ifx\csname natexlab\endcsname\relax\def\natexlab#1{#1}\fi

\bibitem[{Arcadinho et~al.(2022)Arcadinho, Aparício, Veiga, and Alegria}]{Arcadinho2022T5QL}
Samuel Arcadinho, David Aparício, Hugo Veiga, and António Alegria. 2022.
\newblock \href {http://arxiv.org/abs/2209.10254} {{{T5QL}}: {{Taming}} language models for {{SQL}} generation}.

\bibitem[{Bogin et~al.(2019)Bogin, Gardner, and Berant}]{Bogin2019Global}
Ben Bogin, Matt Gardner, and Jonathan Berant. 2019.
\newblock \href {https://doi.org/10.18653/v1/D19-1378} {Global {{Reasoning}} over {{Database Structures}} for {{Text-to-SQL Parsing}}}.
\newblock In \emph{Proceedings of the 2019 {{Conference}} on {{Empirical Methods}} in {{Natural Language Processing}} and the 9th {{International Joint Conference}} on {{Natural Language Processing}} ({{EMNLP-IJCNLP}})}, pages 3659--3664, {Hong Kong, China}. {Association for Computational Linguistics}.

\bibitem[{Brody et~al.(2022)Brody, Alon, and Yahav}]{Brody2022How}
Shaked Brody, Uri Alon, and Eran Yahav. 2022.
\newblock \href {https://openreview.net/forum?id=F72ximsx7C1} {How attentive are graph attention networks?}
\newblock In \emph{International Conference on Learning Representations}.

\bibitem[{Cao et~al.(2021)Cao, Chen, Chen, Zhao, Zhu, and Yu}]{Cao2021LGESQL}
Ruisheng Cao, Lu~Chen, Zhi Chen, Yanbin Zhao, Su~Zhu, and Kai Yu. 2021.
\newblock \href {https://doi.org/10.18653/v1/2021.acl-long.198} {{{LGESQL}}: {{Line Graph Enhanced Text-to-SQL Model}} with {{Mixed Local}} and {{Non-Local Relations}}}.
\newblock In \emph{Proceedings of the 59th {{Annual Meeting}} of the {{Association}} for {{Computational Linguistics}} and the 11th {{International Joint Conference}} on {{Natural Language Processing}} ({{Volume}} 1: {{Long Papers}})}, pages 2541--2555, {Online}. {Association for Computational Linguistics}.

\bibitem[{Center(1987)}]{OhioSupercomputerCenter1987}
Ohio~Supercomputer Center. 1987.
\newblock \href {http://osc.edu/ark:/19495/f5s1ph73} {Ohio supercomputer center}.

\bibitem[{Chen et~al.(2023)Chen, Chen, White, Mooney, Payani, Srinivasa, Su, and Sun}]{chen-etal-2023-sqledit}
Ziru Chen, Shijie Chen, Michael White, Raymond Mooney, Ali Payani, Jayanth Srinivasa, Yu~Su, and Huan Sun. 2023.
\newblock \href {https://doi.org/10.18653/v1/2023.acl-short.117} {Text-to-{SQL} error correction with language models of code}.
\newblock In \emph{Proceedings of the 61st Annual Meeting of the Association for Computational Linguistics (Volume 2: Short Papers)}, pages 1359--1372, Toronto, Canada. Association for Computational Linguistics.

\bibitem[{Feng et~al.(2020)Feng, Guo, Tang, Duan, Feng, Gong, Shou, Qin, Liu, Jiang, and Zhou}]{Feng2020CodeBERT}
Zhangyin Feng, Daya Guo, Duyu Tang, Nan Duan, Xiaocheng Feng, Ming Gong, Linjun Shou, Bing Qin, Ting Liu, Daxin Jiang, and Ming Zhou. 2020.
\newblock \href {https://doi.org/10.18653/v1/2020.findings-emnlp.139} {{{CodeBERT}}: {{A Pre-Trained Model}} for {{Programming}} and {{Natural Languages}}}.
\newblock In \emph{Findings of the {{Association}} for {{Computational Linguistics}}: {{EMNLP}} 2020}, pages 1536--1547, {Online}. {Association for Computational Linguistics}.

\bibitem[{Gan et~al.(2021)Gan, Chen, Xie, Purver, Woodward, Drake, and Zhang}]{Gan2021Natural}
Yujian Gan, Xinyun Chen, Jinxia Xie, Matthew Purver, John~R. Woodward, John Drake, and Qiaofu Zhang. 2021.
\newblock \href {https://doi.org/10.18653/v1/2021.findings-emnlp.174} {Natural {{SQL}}: {{Making SQL Easier}} to {{Infer}} from {{Natural Language Specifications}}}.
\newblock In \emph{Findings of the {{Association}} for {{Computational Linguistics}}: {{EMNLP}} 2021}, pages 2030--2042, {Punta Cana, Dominican Republic}. {Association for Computational Linguistics}.

\bibitem[{Guo et~al.(2019)Guo, Zhan, Gao, Xiao, Lou, Liu, and Zhang}]{Guo2019Towards}
Jiaqi Guo, Zecheng Zhan, Yan Gao, Yan Xiao, Jian-Guang Lou, Ting Liu, and Dongmei Zhang. 2019.
\newblock \href {https://doi.org/10.18653/v1/P19-1444} {Towards complex text-to-{SQL} in cross-domain database with intermediate representation}.
\newblock In \emph{Proceedings of the 57th Annual Meeting of the Association for Computational Linguistics}, pages 4524--4535, Florence, Italy. Association for Computational Linguistics.

\bibitem[{Gur et~al.(2018)Gur, Yavuz, Su, and Yan}]{Gur2018DialSQL}
Izzeddin Gur, Semih Yavuz, Yu~Su, and Xifeng Yan. 2018.
\newblock \href {https://doi.org/10.18653/v1/P18-1124} {{{DialSQL}}: {{Dialogue Based Structured Query Generation}}}.
\newblock In \emph{Proceedings of the 56th {{Annual Meeting}} of the {{Association}} for {{Computational Linguistics}} ({{Volume}} 1: {{Long Papers}})}, pages 1339--1349, {Melbourne, Australia}. {Association for Computational Linguistics}.

\bibitem[{Jorgensen and Shepperd(2007)}]{bugfixhard1}
Magne Jorgensen and Martin Shepperd. 2007.
\newblock \href {https://doi.org/10.1109/TSE.2007.256943} {A systematic review of software development cost estimation studies}.
\newblock \emph{IEEE Transactions on Software Engineering}, 33(1):33--53.

\bibitem[{Kelkar et~al.(2020)Kelkar, Relan, Bhardwaj, Vaichal, Khatri, and Relan}]{Kelkar2020BertrandDR}
Amol Kelkar, Rohan Relan, Vaishali Bhardwaj, Saurabh Vaichal, Chandra Khatri, and Peter Relan. 2020.
\newblock \href {https://doi.org/10.48550/ARXIV.2002.00557} {Bertrand-dr: Improving text-to-sql using a discriminative re-ranker}.

\bibitem[{Lee et~al.(2021)Lee, Polozov, and Richardson}]{lpr2021kaggledbqa}
Chia-Hsuan Lee, Oleksandr Polozov, and Matthew Richardson. 2021.
\newblock \href {https://doi.org/10.18653/v1/2021.acl-long.176} {{K}aggle{DBQA}: Realistic evaluation of text-to-{SQL} parsers}.
\newblock In \emph{Proceedings of the 59th Annual Meeting of the Association for Computational Linguistics and the 11th International Joint Conference on Natural Language Processing (Volume 1: Long Papers)}, pages 2261--2273, Online. Association for Computational Linguistics.

\bibitem[{Li et~al.(2023)Li, Zhang, Li, and Chen}]{li2022resdsql}
Haoyang Li, Jing Zhang, Cuiping Li, and Hong Chen. 2023.
\newblock Resdsql: Decoupling schema linking and skeleton parsing for text-to-sql.
\newblock In \emph{AAAI}.

\bibitem[{Li et~al.(2020)Li, Chen, Liu, Gao, Lou, Zhang, and Zhang}]{Li2020What}
Yuntao Li, Bei Chen, Qian Liu, Yan Gao, Jian-Guang Lou, Yan Zhang, and Dongmei Zhang. 2020.
\newblock \href {https://doi.org/10.18653/v1/2020.emnlp-main.561} {``{{What Do You Mean}} by {{That}}?'' {{A Parser-Independent Interactive Approach}} for {{Enhancing Text-to-SQL}}}.
\newblock In \emph{Proceedings of the 2020 {{Conference}} on {{Empirical Methods}} in {{Natural Language Processing}} ({{EMNLP}})}, pages 6913--6922, {Online}. {Association for Computational Linguistics}.

\bibitem[{Lin et~al.(2020)Lin, Socher, and Xiong}]{Lin2020Bridging}
Xi~Victoria Lin, Richard Socher, and Caiming Xiong. 2020.
\newblock \href {https://doi.org/10.18653/v1/2020.findings-emnlp.438} {Bridging {{Textual}} and {{Tabular Data}} for {{Cross-Domain Text-to-SQL Semantic Parsing}}}.
\newblock In \emph{Findings of the {{Association}} for {{Computational Linguistics}}: {{EMNLP}} 2020}, pages 4870--4888, {Online}. {Association for Computational Linguistics}.

\bibitem[{Loshchilov and Hutter(2019)}]{Loshchilov2019Decoupled}
Ilya Loshchilov and Frank Hutter. 2019.
\newblock Decoupled weight decay regularization.
\newblock In \emph{International Conference on Learning Representations}.

\bibitem[{Pi et~al.(2022)Pi, Wang, Gao, Guo, Li, and Lou}]{Pi2022Towards}
Xinyu Pi, Bing Wang, Yan Gao, Jiaqi Guo, Zhoujun Li, and Jian-Guang Lou. 2022.
\newblock \href {https://doi.org/10.18653/v1/2022.acl-long.142} {Towards robustness of text-to-{SQL} models against natural and realistic adversarial table perturbation}.
\newblock In \emph{Proceedings of the 60th Annual Meeting of the Association for Computational Linguistics (Volume 1: Long Papers)}, pages 2007--2022, Dublin, Ireland. Association for Computational Linguistics.

\bibitem[{Qi et~al.(2022)Qi, Tang, He, Wan, Cheng, Zhou, Wang, Zhang, and Lin}]{qi-etal-2022-rasat}
Jiexing Qi, Jingyao Tang, Ziwei He, Xiangpeng Wan, Yu~Cheng, Chenghu Zhou, Xinbing Wang, Quanshi Zhang, and Zhouhan Lin. 2022.
\newblock \href {https://aclanthology.org/2022.emnlp-main.211} {{RASAT}: Integrating relational structures into pretrained {S}eq2{S}eq model for text-to-{SQL}}.
\newblock In \emph{Proceedings of the 2022 Conference on Empirical Methods in Natural Language Processing}, pages 3215--3229, Abu Dhabi, United Arab Emirates. Association for Computational Linguistics.

\bibitem[{Qi et~al.(2020)Qi, Zhang, Zhang, Bolton, and Manning}]{Qi2020Stanza}
Peng Qi, Yuhao Zhang, Yuhui Zhang, Jason Bolton, and Christopher~D. Manning. 2020.
\newblock \href {https://doi.org/10.18653/v1/2020.acl-demos.14} {Stanza: {{A Python Natural Language Processing Toolkit}} for {{Many Human Languages}}}.
\newblock In \emph{Proceedings of the 58th {{Annual Meeting}} of the {{Association}} for {{Computational Linguistics}}: {{System Demonstrations}}}, pages 101--108, {Online}. {Association for Computational Linguistics}.

\bibitem[{Rubin and Berant(2021)}]{Rubin2021SmBoP}
Ohad Rubin and Jonathan Berant. 2021.
\newblock \href {https://doi.org/10.18653/v1/2021.naacl-main.29} {{{SmBoP}}: {{Semi-autoregressive Bottom-up Semantic Parsing}}}.
\newblock In \emph{Proceedings of the 2021 {{Conference}} of the {{North American Chapter}} of the {{Association}} for {{Computational Linguistics}}: {{Human Language Technologies}}}, pages 311--324, {Online}. {Association for Computational Linguistics}.

\bibitem[{Scholak et~al.(2021)Scholak, Schucher, and Bahdanau}]{Scholak2021PICARD}
Torsten Scholak, Nathan Schucher, and Dzmitry Bahdanau. 2021.
\newblock \href {https://doi.org/10.48550/ARXIV.2109.05093} {Picard: Parsing incrementally for constrained auto-regressive decoding from language models}.

\bibitem[{Wang et~al.(2020)Wang, Shin, Liu, Polozov, and Richardson}]{Wang2020RATSQLa}
Bailin Wang, Richard Shin, Xiaodong Liu, Oleksandr Polozov, and Matthew Richardson. 2020.
\newblock \href {https://doi.org/10.18653/v1/2020.acl-main.677} {{{RAT-SQL}}: {{Relation-Aware Schema Encoding}} and {{Linking}} for {{Text-to-SQL Parsers}}}.
\newblock In \emph{Proceedings of the 58th {{Annual Meeting}} of the {{Association}} for {{Computational Linguistics}}}, pages 7567--7578, {Online}. {Association for Computational Linguistics}.

\bibitem[{Weiss et~al.(2007)Weiss, Premraj, Zimmermann, and Zeller}]{bugfixhard2}
Cathrin Weiss, Rahul Premraj, Thomas Zimmermann, and Andreas Zeller. 2007.
\newblock \href {https://doi.org/10.1109/MSR.2007.13} {How long will it take to fix this bug?}
\newblock In \emph{Fourth International Workshop on Mining Software Repositories (MSR'07:ICSE Workshops 2007)}, pages 1--1.

\bibitem[{Yao et~al.(2019)Yao, Su, Sun, and Yih}]{Yao2019Modelbased}
Ziyu Yao, Yu~Su, Huan Sun, and Wen-tau Yih. 2019.
\newblock \href {https://doi.org/10.18653/v1/D19-1547} {Model-based {{Interactive Semantic Parsing}}: {{A Unified Framework}} and {{A Text-to-SQL Case Study}}}.
\newblock In \emph{Proceedings of the 2019 {{Conference}} on {{Empirical Methods}} in {{Natural Language Processing}} and the 9th {{International Joint Conference}} on {{Natural Language Processing}} ({{EMNLP-IJCNLP}})}, pages 5447--5458, {Hong Kong, China}. {Association for Computational Linguistics}.

\bibitem[{Yao et~al.(2020)Yao, Tang, Yih, Sun, and Su}]{Yao2020Imitation}
Ziyu Yao, Yiqi Tang, Wen-tau Yih, Huan Sun, and Yu~Su. 2020.
\newblock \href {https://doi.org/10.18653/v1/2020.emnlp-main.559} {An {{Imitation Game}} for {{Learning Semantic Parsers}} from {{User Interaction}}}.
\newblock In \emph{Proceedings of the 2020 {{Conference}} on {{Empirical Methods}} in {{Natural Language Processing}} ({{EMNLP}})}, pages 6883--6902, {Online}. {Association for Computational Linguistics}.

\bibitem[{Yin and Neubig(2017)}]{Yin2017Syntactica}
Pengcheng Yin and Graham Neubig. 2017.
\newblock \href {https://doi.org/10.18653/v1/P17-1041} {A {{Syntactic Neural Model}} for {{General-Purpose Code Generation}}}.
\newblock In \emph{Proceedings of the 55th {{Annual Meeting}} of the {{Association}} for {{Computational Linguistics}} ({{Volume}} 1: {{Long Papers}})}, pages 440--450. {Association for Computational Linguistics}.

\bibitem[{Yin and Neubig(2019)}]{yin-neubig-2019-reranking}
Pengcheng Yin and Graham Neubig. 2019.
\newblock \href {https://doi.org/10.18653/v1/P19-1447} {Reranking for neural semantic parsing}.
\newblock In \emph{Proceedings of the 57th Annual Meeting of the Association for Computational Linguistics}, pages 4553--4559, Florence, Italy. Association for Computational Linguistics.

\bibitem[{Yu et~al.(2021)Yu, Wu, Lin, Wang, Tan, Yang, Radev, Socher, and Xiong}]{Yu2021Grappa}
Tao Yu, Chien-Sheng Wu, Xi~Victoria Lin, Bailin Wang, Yi~Chern Tan, Xinyi Yang, Dragomir Radev, Richard Socher, and Caiming Xiong. 2021.
\newblock \href {https://arxiv.org/abs/2009.13845} {Grappa: Grammar-augmented pre-training for table semantic parsing}.
\newblock In \emph{International Conference on Learning Representations}.

\bibitem[{Yu et~al.(2018)Yu, Zhang, Yang, Yasunaga, Wang, Li, Ma, Li, Yao, Roman, Zhang, and Radev}]{Yu2018Spider}
Tao Yu, Rui Zhang, Kai Yang, Michihiro Yasunaga, Dongxu Wang, Zifan Li, James Ma, Irene Li, Qingning Yao, Shanelle Roman, Zilin Zhang, and Dragomir Radev. 2018.
\newblock \href {https://doi.org/10.18653/v1/D18-1425} {Spider: {{A Large-Scale Human-Labeled Dataset}} for {{Complex}} and {{Cross-Domain Semantic Parsing}} and {{Text-to-SQL Task}}}.
\newblock In \emph{Proceedings of the 2018 {{Conference}} on {{Empirical Methods}} in {{Natural Language Processing}}}, pages 3911--3921, {Brussels, Belgium}. {Association for Computational Linguistics}.

\bibitem[{Zeng et~al.(2020)Zeng, Lin, Hoi, Socher, Xiong, Lyu, and King}]{Zeng2020Photon}
Jichuan Zeng, Xi~Victoria Lin, Steven~C.H. Hoi, Richard Socher, Caiming Xiong, Michael Lyu, and Irwin King. 2020.
\newblock \href {https://doi.org/10.18653/v1/2020.acl-demos.24} {Photon: {{A Robust Cross-Domain Text-to-SQL System}}}.
\newblock In \emph{Proceedings of the 58th {{Annual Meeting}} of the {{Association}} for {{Computational Linguistics}}: {{System Demonstrations}}}, pages 204--214, {Online}. {Association for Computational Linguistics}.

\end{thebibliography}
\bibliographystyle{acl_natbib}

\appendix
\setcounter{table}{0}
\renewcommand\thetable{\Alph{section}.\arabic{table}}
\setcounter{figure}{0}
\renewcommand\thefigure{\Alph{section}.\arabic{figure}}
\section*{Appendices}
We provide additional details as follows:
\begin{enumerate}
    \item Appendix \ref{sec:appendix_fixes_spider}: Fixed Spider Evaluation Script    
    \item Appendix \ref{sec:appendix_sqlite_grammar}: Modified SQLite Grammar
    \item Appendix \ref{sec:appendix_beam_examples}: Qualitative Beam Examples 
\end{enumerate}

\section{Fixed Spider Evaluation Script}
\label{sec:appendix_fixes_spider}
We fix the following problems in the official evaluation process of the Spider dataset.
\begin{enumerate}
  \item Incorrect handling of UTF-8 encoded databases. We fix this issue in the Python implementation of the evaluation script. 
  \item Certain examples have gold queries that execute to an empty result, resulting in false positive labels. For these samples, we use exact set match score instead.
  \item Considering the order of returned results when unnecessary. We fix this issue by sorting each column in the execution result in alphabetical order when there is no 'Order By' clause at the top level.
  \item Duplicated values for queries with 'limit' clause. This results in false negatives for questions querying the maximum or minimum values of a column or aggregation. We fix this issue by removing the \texttt{limit} clause in the gold and predicted SQL when their argument is the same. 
\end{enumerate}

We show the accuracy of each parser on Spider's development set based on the official and our version of evaluation in Table \ref{tab:fixed_eval}. Although the differences in accuracy are within 1\%, the disagreements impact 3.3\% to 4\% of the examples. We use the fixed evaluation script to reduce false positive and false negative labels both for training data collection and evaluation of our model.

\begin{table}[htbp]
  \small
  \centering
  \begin{tabular}{cccc}
    \toprule
    Parser & Acc$^*$ & Acc & Disagreement\\
    \midrule
    SmBoP & 75.0 & 75.3 & 41 \\
    \sjc{RESDSQL} & 77.9 & 78.5 & 45 \\
    NatSQL & 71.3 & 71.9 & 40 \\
    \bottomrule
  \end{tabular}
  \caption{\label{tab:fixed_eval} Execution accuracy of the three parsers on Spider's development set using official evaluation script (Acc) and our fixed evaluation script (Acc$^*$). Disagreement counts the number of samples with different labels. }
\end{table}
\begin{table*}[h]
  \small
  \centering
  \begin{tabular}{cccccc}
  \toprule
  Parser & Model & Easy (24.0\%) & Medium (43.1\%) & Hard (16.8\%) & Extra Hard (16.1\%)\\
  \midrule
  \multirow{2}{*}{SmBoP} & CodeBERT &85.0/84.9 &\textbf{81.0}/\textbf{80.5} &75.6/81.8 &63.3/76.4\\
   & CodeBERT+GAT &\textbf{87.1}/\textbf{86.9} & 80.5/78.8 & \textbf{80.6}/\textbf{82.7} & \textbf{65.1}/\textbf{76.6}\\
  \midrule
  \multirow{2}{*}{\sjc{RESDSQL}} & CodeBERT & 91.1/86.9 & 81.4/\textbf{77.5} & 75.5/\textbf{81.1} & 65.0/\textbf{76.3}\\
   & CodeBERT+GAT & \textbf{91.3}/\textbf{91.1} & \textbf{83.3}/76.9 & \textbf{76.0}/80.9 & \textbf{66.4}/75.9\\
  \midrule
  \multirow{2}{*}{NatSQL} & CodeBERT & \textbf{89.0}/\textbf{91.9} & \textbf{82.6}/\textbf{84.6} & 78.5/84.6 & 72.1/78.5\\
   & CodeBERT+GAT & 88.7/91.6 & 81.4/84.0 & \textbf{78.9}/\textbf{87.3} & \textbf{74.1}/\textbf{80.1}\\
  \bottomrule
  \end{tabular}
  \caption{\label{tab:perf_difficulty} Error detection performance Acc/AUC break down by difficulty on the Spider dev set. Difficulty is decided by the official Spider evaluation script. The proportion of each difficulty type is in the parenthesis.}
\end{table*}

\section{KaggleDBQA Results}
\label{sec:appendix_kaggledbqa}

\begin{table}[htbp]
  \small
  \centering
  \begin{tabular}{lc}
    \toprule
    Re-ranker & CodeBERT+GAT\\
    \midrule
    N/A & 20.5 \\
    \midrule
    RR & \textbf{21.8} \\
    ED + RR & 21.4\\
    \midrule
    Beam Hit & 25.9 \\
    \bottomrule
  \end{tabular}
  \caption{\label{tab:reranking_kaggle} Execution accuracy of BRIDGE v2 and re-ranking using the CodeBERT + GAT model (left) and CodeBERT (right) on KaggleDBQA. RR: Re-ranking all beams; ED+RR: Re-ranking beams after error detection.}
\end{table}

To test the generalization abilities of the proposed error detecor, we perform 0-shot evaluations on the 370 test examples\footnote{We use all examples in \lstinline{examples/*_test.json} in \url{https://github.com/chiahsuan156/KaggleDBQA}} in KaggleDBQA\cite{lpr2021kaggledbqa}. KaggleDBQA features more realistic database naming and makes entity linking significantly more challenging than Spider. We only experiment with BRIDGE v2 for the following reasons: (1) Under the 0-shot testing, SmBoP trained on Spider got an accuracy of 1.6\% both for top-1 and beam hit, making re-ranking meaningless. This is partly due to the failure of its entity linking modules based on span extraction from questions. (2) At the time of writing, the SQL-to-NatSQL part of NatSQL's preprocessing code has not been released, and its current codebase does not support KaggleDBQA. \sjc{The same applies for RESDSQL-large + NatSQL.}

We present the 0-shot re-ranking results with BRIDGE v2 using CodeBERT+GAT in Table \ref{tab:reranking_kaggle}. Without any training data, CodeBERT+GAT improves BRIDGE v2's accuracy by 1.3\%.

\section{Performance by Difficulty}
\label{sec:appendix_perf_difficulty}

In Table \ref{tab:perf_difficulty}, we break down the error detection performance in Table \ref{tab:ed_perf} by question difficulty and compare the performance of CodeBERT and CodeBERT+GAT. While CodeBERT can perform better on \textit{easy} and \textit{medium} questions for some parsers, CodeBERT+GAT consistently wins on \textit{hard} and \textit{extra hard} questions, showing the effectiveness of introducing structural features for harder questions. Since 63.1\% of the questions are of \textit{easy} or \textit{medium} difficulty, the overall evaluation in Table \ref{tab:ed_perf} favors CodeBERT.

\section{Modified SQLite Grammar}
\label{sec:appendix_sqlite_grammar}
we use a modified version based on a publicly available context-free grammar for SQLite \url{https://github.com/antlr/grammars-v4/tree/master/sql/sqlite}, 
We slightly modify the publicly available SQLite grammar for Antlr4\footnote{\url{https://github.com/antlr/grammars-v4/tree/master/sql/sqlite}} to introduce more hierarchical structures of SQL queries at the top level.
\paragraph*{Terminals} We represent SQL keyword 'GROUP BY' by a single terminal \texttt{GROUP\_BY\_} and 'ORDER BY' by \texttt{ORDER\_BY\_}. The original grammar reuses \texttt{BY\_} for 'BY', which we think breaks the entirety of these two keywords.

\paragraph{Non-terminals} We first remove \texttt{values\_clause} and rules related to window functions, as they are not used by SQL queries in the Spider dataset. 
Then we break the \texttt{select\_core} non-terminal, which represents a SQL query starting with SELECT, into a composition of multiple non-terminals, one for each SQL clause. 

Our new SQLite grammar is listed as follows:
\begin{lstlisting}
  select_core:
    (
      SELECT_ (DISTINCT_ | ALL_)? 
      result_clause 
      (from_clause)? 
      (where_clause)? 
      (group_by_clause)? 
    )
  ;

  result_clause:
    result_column (COMMA result_column)*
  ;

  from_clause:
    FROM_ table_or_subquery 
    (COMMA table_or_subquery)* 
    | FROM_ join_clause
      
  ;

  where_clause:
    WHERE_ expr
  ;
  
  group_by_clause:
    GROUP_ BY_ expr 
    (COMMA expr)* 
    (HAVING_ expr)?
  ;
\end{lstlisting}

The original grammar for \texttt{select\_core}:
\begin{lstlisting}
  select_core:
    (
      SELECT_ (DISTINCT_ | ALL_)? 
      result_column 
      (COMMA result_column)* 
      (FROM_ 
        (table_or_subquery 
          (COMMA table_or_subquery)* 
          | join_clause
        )
      )? 
      (WHERE_ whereExpr=expr)? 
      (GROUP_ BY_ groupByExpr+=expr 
        (COMMA groupByExpr+=expr)* 
        (HAVING_ havingExpr=expr)?
      )?
    )
  ; 
\end{lstlisting}

Notice the excessive use of \texttt{*} in the original grammar that fails to represent the hierarchical relationship between the SELECT statement and each clause.

\section{Qualitative Beam Examples}
\label{sec:appendix_beam_examples}

\begin{table*}[h!]
  \small
  \begin{tabular}{p{0.1\linewidth}p{0.85\linewidth}}
  \toprule
  Question: & How many heads of the departments are older than 56?\\
  Gold SQL: & \texttt{SELECT COUNT(*) FROM head WHERE head.age > 56}\\
  \cmidrule(lr){1-2}
  \multirow{5}{*}{SmBoP}&\texttt{SELECT COUNT(*) FROM head WHERE head.age > 56 }\\
  &\texttt{SELECT head.name FROM head WHERE head.age > 56 }\\
  &\texttt{SELECT MAX(head.age) FROM head WHERE head.age > 56 }\\
  &\texttt{SELECT head.age FROM head WHERE head.age > 56 }\\
  &\texttt{SELECT * FROM head WHERE head.age > 56 }\\
  \cmidrule(lr){1-2}
\multirow{5}{*}{RESDSQL}&\texttt{SELECT COUNT(*) FROM head WHERE head.age > 56 }\\
&\texttt{SELECT COUNT(DISTINCT head.name) FROM head WHERE head.age > 56 }\\
&\texttt{SELECT COUNT(head.head\_id) FROM head WHERE head.age > 56 }\\
&\texttt{SELECT COUNT(*) , department.name FROM management JOIN head ON management.head\_ID = head.head\_ID JOIN department ON management.department\_ID = department.Department\_ID WHERE  head.age > 56 GROUP BY department.name}\\
&\texttt{SELECT ( DISTINCT department.department\_id) from management JOIN head ON management.head\_ID = head.head\_ID JOIN department ON management.department\_ID = department.Department\_ID where  head.age > 56}\\
  \cmidrule(lr){1-2}
\multirow{5}{*}{NatSQL}&\texttt{SELECT COUNT(*) FROM head WHERE head.age > 56 }\\
  &\texttt{SELECT COUNT(*) FROM department WHERE department.department\_id in (SELECT management.department\_ID FROM management, head WHERE head.age = 56) }\\
  &\texttt{SELECT COUNT(*) FROM head WHERE head.age = 56 }\\
  &\texttt{SELECT COUNT(*) FROM head WHERE head.age < 56 }\\
  &\texttt{SELECT COUNT(*) FROM head WHERE head.age >= 56 }\\
\midrule
  Question: &Show the names of the three most recent festivals.\\  
  Gold SQL: & \texttt{SELECT festival\_detail.festival\_name FROM festival\_detail ORDER BY festival\_detail.year DESC LIMIT 3}\\

  \cmidrule(lr){1-2}
  \multirow{10}{*}{SmBoP}&\texttt{SELECT festival\_detail.festival\_name FROM festival\_detail ORDER BY festival\_detail.year DESC LIMIT 3 }\\
  &\texttt{SELECT festival\_detail.festival\_name FROM festival\_detail WHERE festival\_detail.year = (SELECT MAX( festival\_detail.year ) FROM festival\_detail) }\\
  &\texttt{SELECT 3 FROM festival\_detail WHERE festival\_detail.year = (SELECT MAX( festival\_detail.year ) FROM festival\_detail)}\\
  &\texttt{SELECT MAX( festival\_detail.year ) FROM festival\_detail ORDER BY festival\_detail.year DESC LIMIT 3 }\\
  &\texttt{SELECT MAX( festival\_detail.year ) FROM festival\_detail ORDER BY festival\_detail.year DESC }\\
  \cmidrule(lr){1-2}
\multirow{6}{*}{RESDSQL}&\texttt{SELECT festival\_detail.festival\_name FROM festival\_detail ORDER BY festival\_detail.year DESC LIMIT 3}\\
&\texttt{SELECT festival\_detail.festival\_name FROM festival\_detail ORDER BY festival\_detail.year ASC LIMIT 3}\\
&\texttt{SELECT DISTINCT festival\_detail.festival\_name FROM festival\_detail ORDER BY festival\_detail.year DESC LIMIT 3}\\
  \cmidrule(lr){1-2}
\multirow{10}{*}{NatSQL}&\texttt{SELECT festival\_detail.festival\_name FROM festival\_detail  ORDER BY festival\_detail.year DESC LIMIT 3 }\\
  &\texttt{SELECT festival\_detail.festival\_name FROM festival\_detail ORDER BY festival\_detail.year ASC LIMIT 3  }\\
  &\texttt{SELECT festival\_detail.festival\_name , festival\_detail.year FROM festival\_detail ORDER BY festival\_detail.year DESC LIMIT 3  }\\
  &\texttt{SELECT festival\_detail.festival\_name FROM festival\_detail  }\\
  &\texttt{SELECT festival\_detail.festival\_name FROM festival\_detail GROUP BY festival\_detail.festival\_name  ORDER BY festival\_detail.year DESC LIMIT 3 }\\
\bottomrule
\end{tabular}
\caption{\label{tab:beam_example} Example beam predictions collected from three base parsers in our training dataset.}
\end{table*}

As mentioned in Section \ref{Sec:data-collection}, the three text-to-SQL parsers behave differently. 
We present their beam predictions on two samples in our training split in Table \ref{tab:beam_example}. 
We can observe that SmBoP and NatSQL could generate more executable SQL queries than \sjc{RESDSQL}. Both SmBoP and NatSQL are capable of generating diverse errors, but \sjc{RESDSQL}'s beam predictions are more likely to share prefixes. 
As an example, SmBoP generates diverse \texttt{SELECT} clauses on both samples, while the \sjc{diversity of} \texttt{SELECT} clauses predicted by \sjc{RESDSQL} and NatSQL \sjc{is lower}.

\end{document}